\documentclass{article}

\usepackage[final]{corl_2020} 

\usepackage[utf8]{inputenc} 
\usepackage[T1]{fontenc}    
\usepackage{hyperref}       
\usepackage{url}            
\usepackage{booktabs}       
\usepackage{amsfonts}       
\usepackage{nicefrac}       
\usepackage{microtype}      
\usepackage{graphicx}
\usepackage{amsmath}
\usepackage{subfigure}
\usepackage{verbatim}
\usepackage{booktabs} 
\usepackage{graphicx,wrapfig,lipsum}
\usepackage{enumitem}
\setlist[itemize]{leftmargin=*}

\usepackage{hyperref}

\title{Augmenting GAIL with BC for sample efficient imitation learning}

\author{
    Rohit Jena  \\
    The Robotics Institute \\
    Carnegie Mellon University \\
    \texttt{rjena@cs.cmu.edu} \\
    \And
    Changliu Liu \\
    The Robotics Institute \\
    Carnegie Mellon University \\
    \texttt{cliu6@andrew.cmu.edu} \\
    \And
    Katia Sycara \\
    The Robotics Institute \\
    Carnegie Mellon University \\
    \texttt{katia@cs.cmu.edu} \\
}

\begin{document}
\maketitle

\begin{abstract}
    Imitation learning is the problem of recovering an expert policy without access to a reward signal.
    Behavior cloning and GAIL are two widely used methods for performing imitation learning.
    Behavior cloning converges in a few iterations, but does not achieve peak performance due to its inherent iid assumption about the state-action distribution.
    GAIL addresses the issue by accounting for the temporal dependencies when performing a state distribution matching between the agent and the expert.
    Although GAIL is sample efficient in the number of expert trajectories required, it is still not very sample efficient in terms of the environment interactions needed for convergence of the policy.
    Given the complementary benefits of both methods, we present a simple and elegant method to combine both methods to enable stable and sample efficient learning.
    Our algorithm is very simple to implement and integrates with different policy gradient algorithms.
    We demonstrate the effectiveness of the algorithm in low dimensional control tasks, gridworlds and in high dimensional image-based tasks.
\end{abstract}

\section{Introduction}
\label{introduction}
We attempt to solve the problem of imitation learning, where a task has to be performed by an agent using only expert demonstrations.
The agent cannot query for more information from the expert in an iterative manner.
One approach to solve the problem is to use behaviour cloning, where learning from demonstrations has been formulated as a supervised learning task.
However, supervised learning assumes the data to be i.i.d. which is an incorrect assumption since the action taken at a state influences the future actions that the expert might take.
Here, the i.i.d. assumption is with respect to the transition function.
In behavior cloning, the assumption is that the action for a given state will not influence the distribution of states that the agent sees later on. Due to this incorrect assumption, behavior cloning cannot deal with covariate shift.
Therefore, the `best' action for a state is chosen from the expert and no other environment dynamics are considered.
However, in RL, the log-probabilities are weighted by the $Q$-function, which encode the dynamics of the transition function and rewards from future states.

This i.i.d. assumption leads to compounding errors in behaviour cloning, and a large number of expert state action pairs must be provided to mitigate this error.
GAIL, on the other hand, is very sample efficient in terms of the number of expert trajectories required but is very sample inefficient in terms of environment interactions.
Environment interactions in many such scenarios require massive amounts of compute time and space.
This is all the more problematic in real-world problems where additional concerns of safety, wear and tear and cost also kick in.
This calls for the need of sample efficient algorithms which require minimal environment interactions.

Off-policy imitation learning may be a viable strategy, however, truly off-policy data is hard to learn from (\citet{trulyoffpolicy}).
Off-policy algorithms are also difficult to implement and often require delicate replay buffer manipulations.
Off-policy may also not be an option when there are ethical or privacy related issues regarding persistent storage of data (EU regulations, for example).
To that end, we propose a strategy to perform imitation learning in an on-policy manner to outperform GAIL in terms of sample efficiency.
Our method uses the fact that behavior cloning is a fast learning procedure but cannot be used as a pre-training step for GAIL as shown in the experiments section.
Our method is simple to implement and empirically we show that our algorithm requires fewer environment interactions across different types of policy gradient algorithms and environments.
Our method can also be extended to an off-policy setting with virtually no modifications to the algorithm.
We demonstrate the versatility of our algorithm on a set of MuJoCo environments, grid world, and image-based Car Racing environments.

\vspace{-6pt}
\section{Related Work}
\vspace{-6pt}
\label{relatedwork}
To mitigate the compounding errors in the naive supervised approach, \citet{forwardtraining} train an iterative algorithm where at each time step t, the policy $\pi_t$ learns the expert behavior on the trajectories induced by $\pi_1 \ldots \pi_{t-1}$.
\citet{forwardtraining} also introduce a stochastic mixing algorithm based on \citet{searn}.
The initial policy starts off as the expert policy, and at each iteration, a new policy is obtained by training on the trajectory induced by the previous policy $\pi_{t-1}$.
The policy at timestep t is obtained by a geometric stochastic mixing of the expert and the previous policies.
\citet{dagger} train a policy using the expert demonstrations, generate new trajectories and use the expert to correct the behavior in these new trajectories iteratively.
Although this method performs much better in a variety of scenarios, it requires access to the expert, which might be very expensive.

Recently, adversarial imitation learning methods have shown successes in a variety of imitation tasks, from low dimensional continous control to high dimensional tasks like autonomous driving from raw pixels as input.
\citet{ho2016generative} propose a framework for directly extracting a policy from trajectories without performing reinforcement learning inside a loop. This approach utilizes a discriminator to distinguish between the state-action pairs induced by the expert and the policy, and the policy uses the output of the discriminator as the reward.
Different approaches build on top of this method, with \citet{infogail} proposing an algorithm that can infer the latent structure of the expert trajectories without explicit supervision.
This approach maximizes a mutual information term between the trajectory and the latent space to capture the variations in the trajectories.
GAIL was further extended by \citet{airl} to produce a scalable inverse reinforcement learning algorithm based on adversarial reward learning. This approach gives a policy as well as a reward function.
These approaches have led to faster imitation learning in both low and high dimensional tasks.

\section{Pre-training in Imitation Learning}
We discuss in the previous section that GAIL and BC offer complementary benefits for imitation learning.
Therefore, a natural question to ask would be if there are any obvious ways combine the two while keeping their respective benefits.
One approach that has found repeated mentions in the literature is pretraining with BC and then finetuning the policy with GAIL.
Although this sounds like a reasonable strategy,our empirical results show that pretraining with behaviour cloning did not help and the agent learns a suboptimal policy as compared to GAIL trained from scratch.
This observation is not uniquely found by us, as demonstrated by \citet{sasaki2018sample}, where they show that GAIL pretrained with behaviour cloning failed to reach optimal performance as compared to GAIL trained from scratch.
The following subsection discusses the effect of warm-started neural networks, and why that may hinder learning in GAIL after pretraining with behavior cloning.

\begin{table*}[htpb]
    \centering
    \resizebox{\textwidth}{!}{
    \begin{tabular}{|c|c|c|c|c|c|} \hline
         & \textbf{Ant} & \textbf{HalfCheetah} & \textbf{Hopper} & \textbf{Reacher} & \textbf{Walker2d} \\ \hline
        \textbf{BC} & $2967.09 \pm 1223.82$ & $-389.14 \pm 1166.19 $ & $1776.43 \pm 858.93 $& $-86.74 \pm 11.25  $& $788.82 \pm 579.34$ \\ \hline
        \textbf{GAIL} & $2732.78 \pm 1107.56$ & $4546.93 \pm 117.10$ & $3035.83 \pm 720.62$ & $-9.78 \pm 2.26$ & $6718.34 \pm 935.17$  \\ \hline
        \textbf{BC+GAIL}  & $1237.88 \pm 725.48$  & $3808.17 \pm 1298.91$ & $14.13 \pm 30.32$  &  $-9.77 \pm 3.04$  & $712.16 \pm 542.22$ \\ \hline
        \textbf{RED} & $-4952.94 \pm 1551.64$  & $-626.47 \pm 384.42$ & $684.52 \pm 478.09$ & $-49.69 \pm 42.90$ & $940.27 \pm 82.59$ \\ \hline
        \textbf{SAIL} & $2750.59 \pm 938.21$ & $\boldsymbol{4584.18 \pm 86.88}$ & $2307.69 \pm 1198.76$ & $-14.31 \pm 11.84$ & $5993.21 \pm 793.99$ \\ \hline
        \textbf{Ours} & $\boldsymbol{3941.69 \pm 944.67}$  & $4558.09 \pm 89.50$  & $\boldsymbol{3554.35 \pm 165.73}$  & $\boldsymbol{-7.98 \pm 2.66}$  & $\boldsymbol{ 6799.93 \pm 387.85}$ \\ \hline
        \textbf{Random} & $-327.04 \pm 790.06$  & $-922.94 \pm 97.30$  & $15.17 \pm 30.58$  & $-136.72 \pm 23.96$  & $-3.03 \pm 4.49$ \\ \hline
        \textbf{Expert} & $4066.96 \pm 695.57$  & $4501.09 \pm 119.37$ & $3593.06 \pm 19.64$ & $-3.92 \pm 1.78$ & $6512.85 \pm 1116.62$ \\ \hline
    \end{tabular}
    }
    \caption{Performance of imitation learning algorithms. For each method, agents are trained with three random seeds. Final performance is measured by averaging across 60 rollouts for each method, 20 rollouts for each seed.}
    \label{tab:bcgail}
\end{table*}
\subsection{Suboptimal performance of warm-started neural networks}
Pre-training networks has shown a number of successes in deep learning, from image classification to natural language inference among others.
The success of pretraining lies in the fact that it can be used on a base model that can be used to finetune later to domain specific tasks with little data.
However, \citet{he2019rethinking} show that random initialization is very robust and performs \textit{no worse} than pretrained networks.
The networks take longer to train than pretrained networks, but their generalization errors are almost always better than that of pretrained networks as shown in their work.
This holds especially true when networks are trained with less data, which is surprising.
\citet{ash2019difficulty} takes this a step further and shows that warm starting a network might lead to poorer generalization although the training losses may be the same.
In the context of imitation learning, behaviour cloning does not train with all the expert trajectories because some validation data is required to prevent overfitting.
GAIL, however, can work with all of the data, and training can stop when the discriminator loss becomes stable or after a fixed number of environment interactions.
Since the policy is warm-started with a fraction of the expert data during behavior cloning, it may lead to an overall poor generalization error when trained on the entire set of trajectories during GAIL training.

\section{Our method}
\label{method}

The motivation for our method is inspired by the fact that optimizing the behavior cloning term alone leads to the agent learning a mapping from states to actions in a few iterations.
However, the i.i.d. supervised training objective does not consider the sequential decision making aspect at all.
Since there is no information about the transition dynamics or the value of following an action at a state (Q-function), behavior cloning would be suboptimal unless a lot of data is provided.
However, even in a limited data setting, behavior cloning can still learn important features that map states to probable actions.
GAIL, on the other hand, is simply reinforcement learning with a learnt reward function which is provided by the discriminator.
However, the rewards provided by the discriminator are not informative in the beginning of the training procedure, and changes along with the policy that adapts to this reward function.
The uninformative rewards do not provide any strong signal that the agent can use to map states to expert actions.
Experiments in Section \ref{bctemp} shows that simply adding a temporal dependency to the behavior cloning term can improve convergence speed over GAIL without even training the discriminator.

Formally, consider the behavior cloning loss which is given by:
$$ \mathcal{L_{BC}} = -E_{\tau_{E}}[\log(\pi(a | s))] $$

In adversarial imitation learning, we also train a discriminator $D$ parameterized by $\omega$.
The discriminator is trained by minimizing the loss
$$\mathcal{L_{D}} = -E_{\tau_E}[\log(D_{\omega}(s, a))] - E_{\tau_\pi}[\log(1 - D_\omega(s, a))]$$

And the policy is trained using a policy gradient algorithm:
$$\mathcal{L_{P}} = - E_{\tau_\pi}[\log(\pi_\theta(a | s)) A_{\omega, \psi}(s, a)]  $$
where the advantage $A$ is estimated using the value network $V_\psi$ and the discriminator $D_\omega$:
\[
A_{\omega, \psi}(s, a) = -\log(1 - D_{\omega}(s, a)) + \gamma \mathbb{E}_{s' \sim T(s'|s, a)}\left[ V_{\psi}(s') \right]  - V_{\psi}(s)
\]
Let the expert trajectories be denoted by a dataset $\mathcal{D}$, where $\mathcal{D} = \{(s_1, a_1), (s_2,a_2), \ldots (s_N, a_N)\} $, containing tuples of states $s_i$ and actions taken by the expert $a_i$.
Let the state-action visitation probability be denoted by $\rho(s, a)$.
The behavior cloning term can also be written as:
\begin{small}
\begin{align*}
    \mathcal{L_{BC}} &= - \sum_{s, a} \rho_E(s, a) \log(\pi(a | s))
    = - \sum_{s, a} \rho_{\pi}(s, a) \left[ \frac{\rho_{E}(s, a)}{\rho_{\pi}(s, a)} \log(\pi(a | s)) \right] = - E_{\tau_{\pi}} \left[ \frac{\rho_{E}(s, a)}{\rho_{\pi}(s, a)} \log(\pi(a | s)) \right]
\end{align*}
\end{small}
This is nothing but a simple manipulation based on importance sampling that allows us to directly add this term to the GAIL term, giving us the final loss function:
$$ \mathcal{L_{BC}}= - E_{\tau_{\pi}} \left[ \left( \frac{\rho_{E}(s, a)}{\rho_{\pi}(s, a)}  + A_{\omega, \phi}(s, a) \right)  \log(\pi(a | s)) \right] $$
The new advantage term inside the RL term intuitively adds an advantage for greedily following the expert action at a given state.
Since the expert is only available indirectly in the form of samples and assuming a deterministic policy, the value $\rho_{E}(s, a)$ can be replaced with a Kronecker delta function
$$\rho_{E}(s, a) = \delta_{\mathcal{D}}(s, a) = \begin{cases}
    1 \quad \text{if $(s, a) \in \mathcal{D}$} \\
    0 \quad \text{otherwise} \\
\end{cases}$$
This leads to a very interesting interpretation of the first advantage term.
If the expert did not perform some action then there is zero advantage in performing that action, but if the expert does perform the action $a$ at state  $s$, the advantage term is  $\frac{1}{\rho_\pi(s, a)}$ which gives more advantage to the agent for following this behavior if the agent does not follow this behavior already.
If the agent takes action $a$ at state  $s$ and so does the expert, then the advantage term is close to 1, which is still positive, but the advantage decreases as the agent starts imitating the expert more precisely. \\
More generally, a weighted sum of the behavior cloning and GAIL term can be used with coefficients $\alpha$ and $1 - \alpha$ with  $\alpha \in [0, 1]$ resulting in the following policy gradient term:
\[
    - E_{\tau_{\pi}} \left[ \left( \alpha\frac{\rho_{E}(s, a)}{\rho_{\pi}(s, a)}  + \left( 1 - \alpha \right) A_{\omega, \phi}(s, a) \right)  \log(\pi(a | s)) \right] = \alpha \mathcal{L}_{BC} + (1-\alpha) \mathcal{L_{P}}
\]
However in a practical scenario, with lack of data, the behavior cloning term may lead to overfitting and choice of $\alpha$ becomes crucial.
Setting $\alpha$ too high can lead to the GAIL term not having enough impact, and setting it too low does not provide the apparent benefit of fast feature learning from the BC term.
However, we observe that $\alpha$ only needs to be high in the initial stages, and can be ignored in the later stages of training since GAIL rewards would become informative then.
Therefore, simulated annealing is a very elegant way to set the value of $\alpha$.

Simulated annealing is used in optimization techniques for approximating the global optimum of a function.
A common use of annealing is done in the learning rate in training of deep neural networks.
\citet{PAN2019} optimize a sequence of gradually improving mosaic functions that approximate the original non-convex objective using an annealing scheme.
\citet{fujimoto2018addressing} use an exponentially moving average of the parameters of the target Q function at each time step.
\citet{jena} use an annealing scheme to stabilize training in the context of semantic segmentation in medical images.
\begin{figure*}[htpb]
    \centering
    \includegraphics[width=0.28\textwidth]{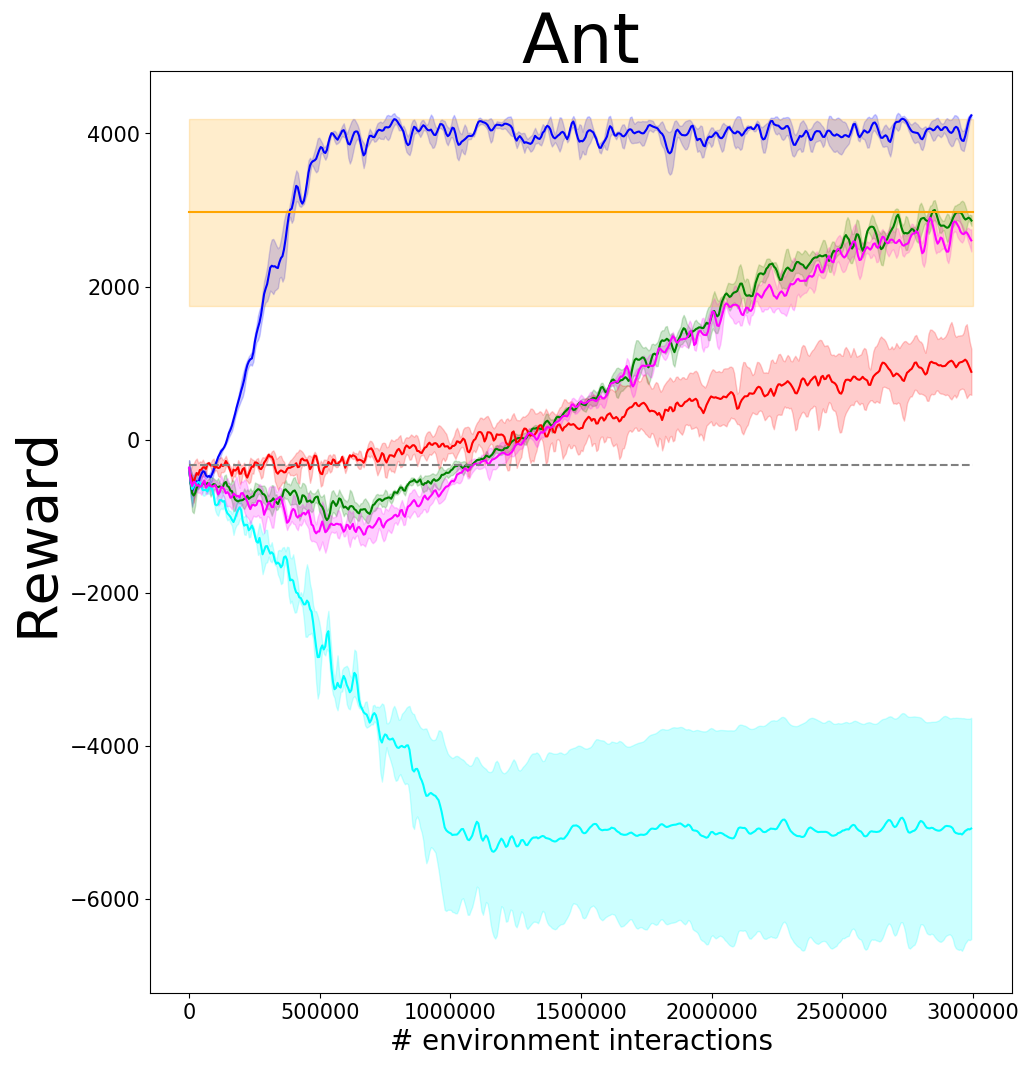}
    \includegraphics[width=0.28\textwidth]{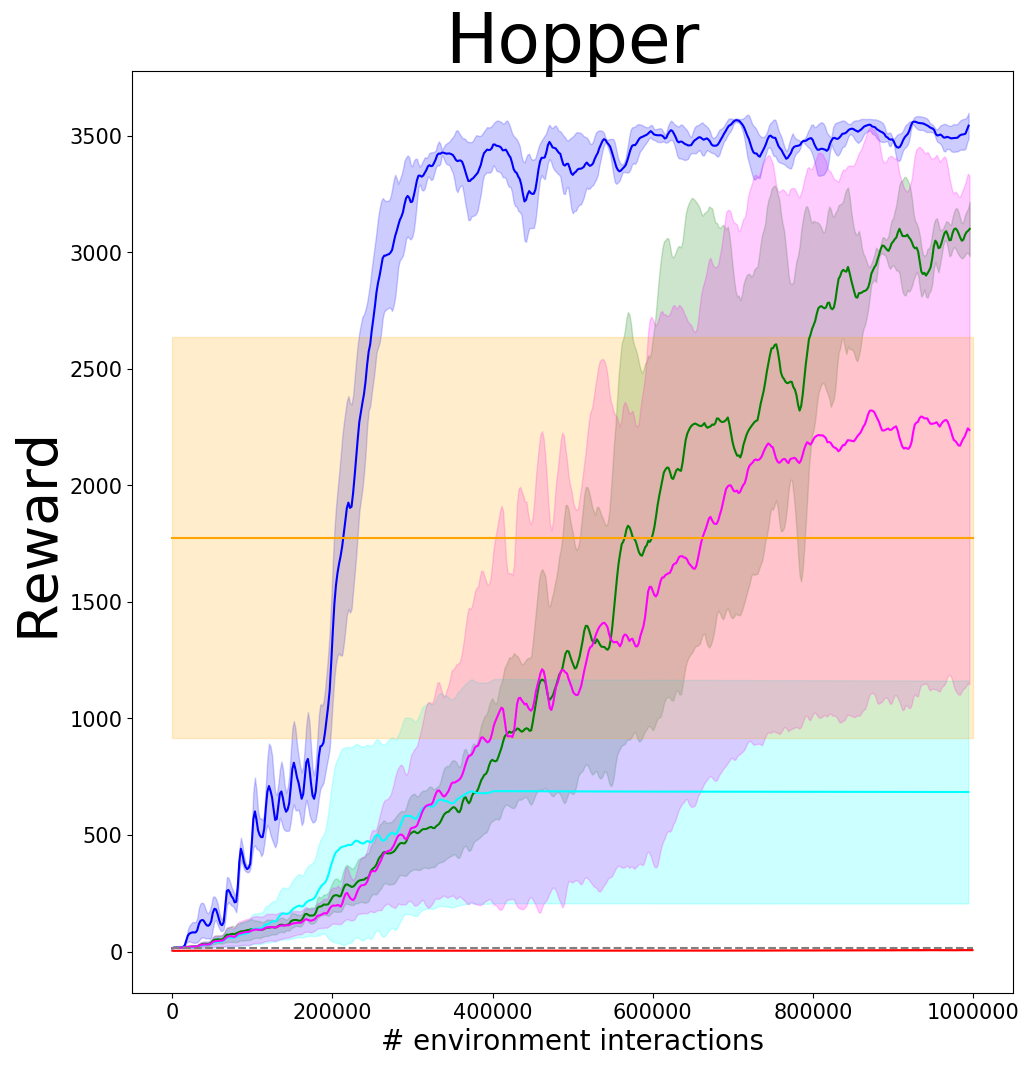}
    \includegraphics[width=0.28\textwidth]{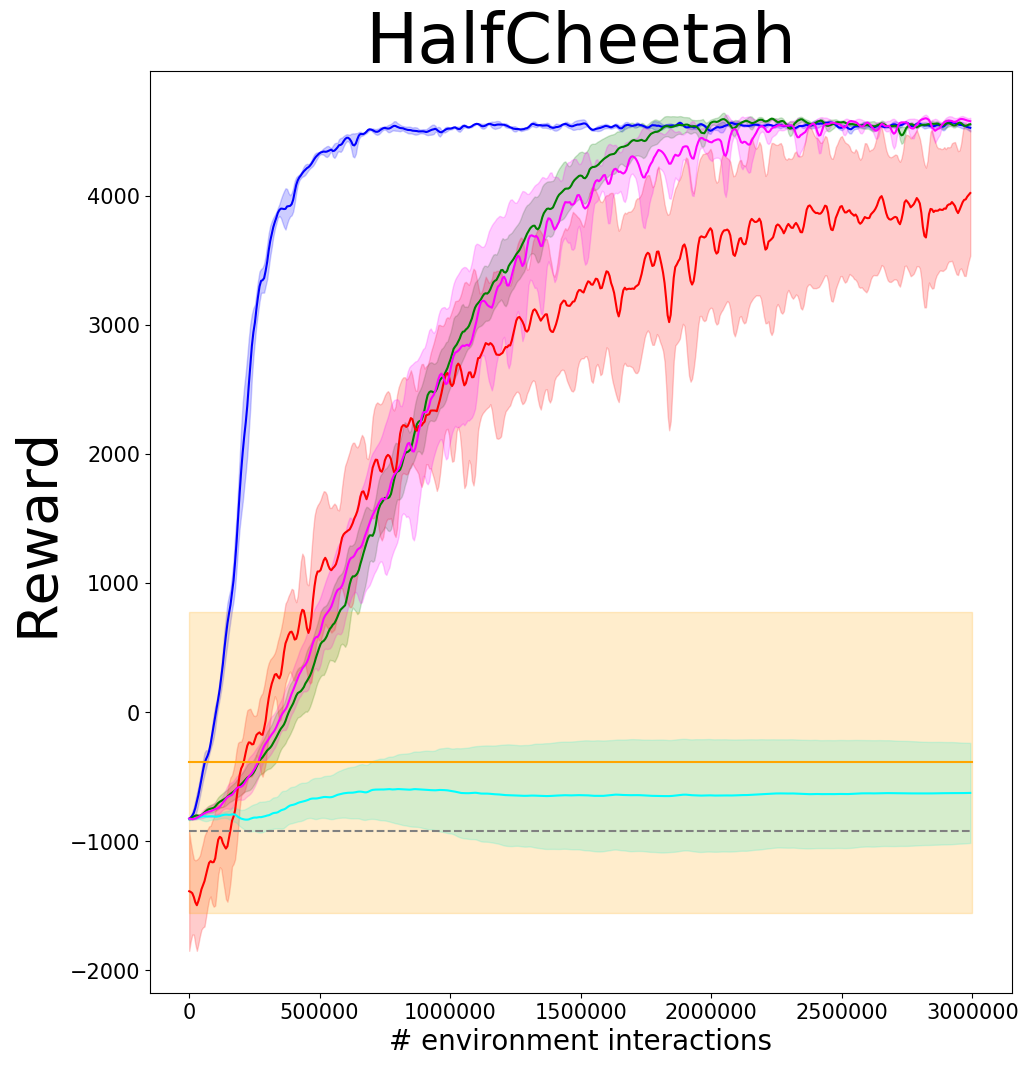}
    \includegraphics[width=0.28\textwidth]{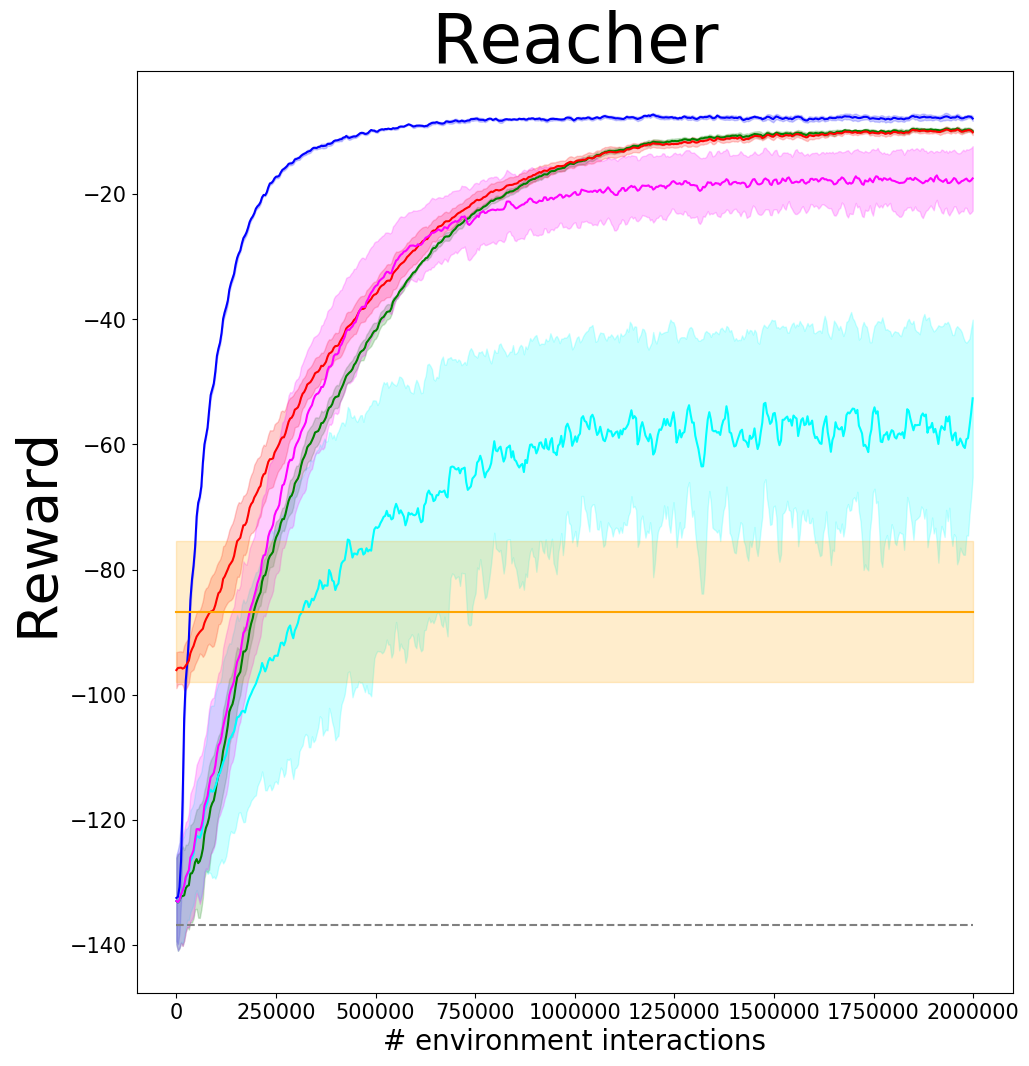}
    \includegraphics[width=0.28\textwidth]{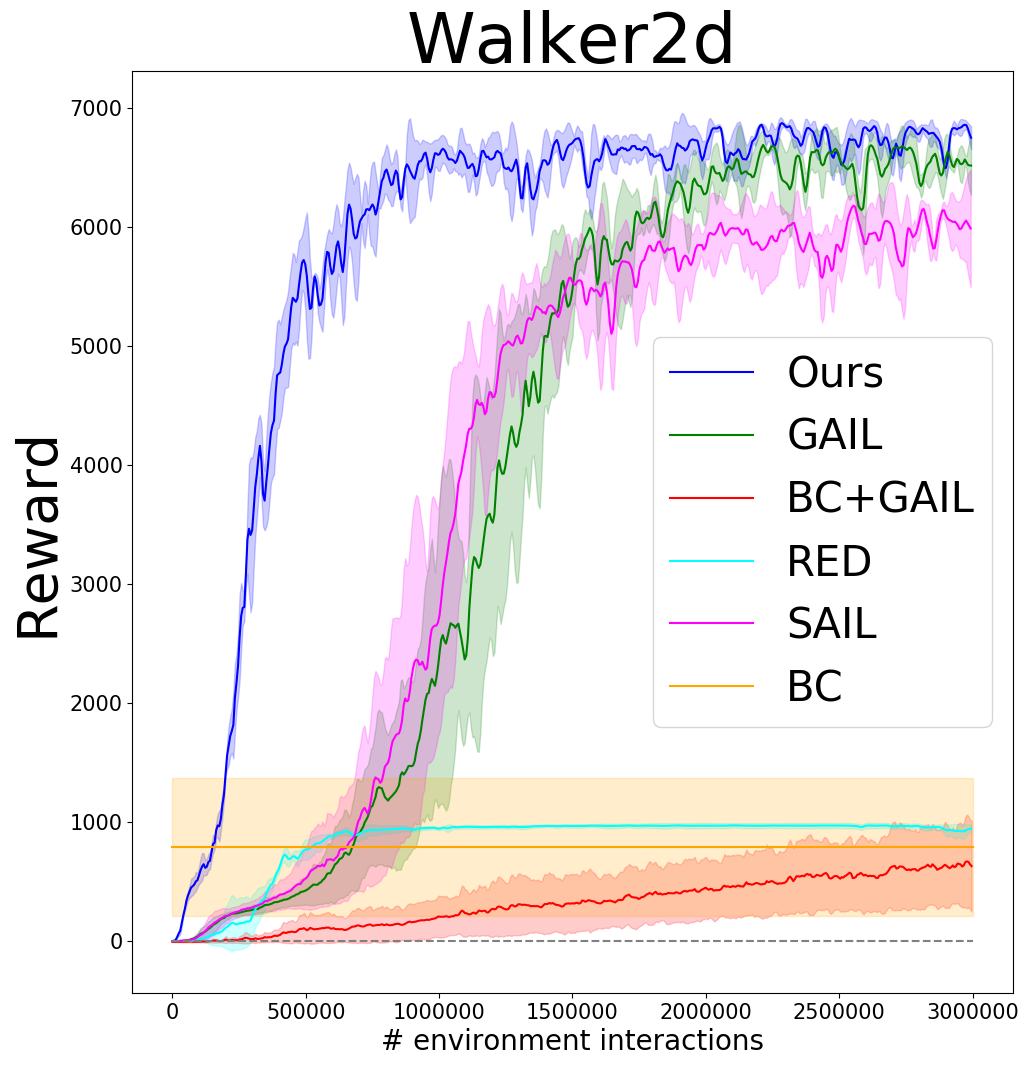}
    \caption{Performance of different imitation learning algorithms on MuJoCo tasks. All methods are tested with 3 random seeds.}
    \label{fig:bcgail}
\end{figure*}

Following these works, we use simulated annealing.
The weighing parameter $\alpha$ is annealed out such that as the number of iterations $t \to \infty$, the optimization looks identical to GAIL, which provides better asymptotic performance.
Specifically, at iteration $t$, we train the policy using the following loss:
$$L^{(t)}_{Total} = \alpha_t \mathcal{L_{BC}} + (1 - \alpha_t)\mathcal{L_P}$$
where $\alpha_t \in [0, 1]$. Note that $\alpha_t = 0$ corresponds to training with GAIL, and $\alpha_t = 1$ corresponds to behaviour cloning.
In our case, we anneal $\alpha_t$ from 1 to 0, which transitions the gradients from a greedy action-matching behaviour to gradually accounting for more long term reward.
The policy produces better-than random behaviour in the initial iterations which provides the discriminator with better trajectories from the policy.
The tradeoff parameter is annealed using an exponential decay $\alpha_t = \alpha_0^{t}$, with $\alpha_0 \in (0, 1)$.

\section{Experiments}
\subsection{Low dimensional control tasks}
We evaluate the proposed algorithm on a variety of continuous control tasks in MuJoCo.
Specifically, we test our method on the \textbf{Ant, Hopper, Half Cheetah, Reacher,} and \textbf{Walker2d} environments.
We compare our algorithm with the following baselines:
\begin{itemize}
\item \textbf{Behavior cloning:} Behavior cloning is a greedy approach to imitation learning.
Although behaviour cloning is very fast since it does not require environment interactions, its asympotic performance is not optimal unless a lot of data is provided.
Since our experiments do not use iterative data collection, we do not use the other behavior cloning baselines which use iterative feedback from experts \citet{forwardtraining}, \citet{dagger}.
\item \textbf{GAIL:} Adversarial imitation learning has been successful in a lot of environments.
    However, adversarial methods are shown to be unstable, and in the presence of low amounts of data, can take a long time to converge.
\item \textbf{BC+GAIL:} \citet{ho2016generative} mention that GAIL can be trained to converge faster by pretraining it with behavior cloning.
    However, they do not report the results for this baseline.
    \citet{sasaki2018sample}, however, report that GAIL pretrained with behavior cloning does not work as effectively as GAIL.
    To make the baseline fairer, we also train the discriminator to differentiate between expert versus pretrained policy trajectories.
\item \textbf{SAIL:} \citet{sail} is the only other method that claims to improve sample efficiency of GAIL without resorting to off-policy methods. Although, our experiments show that the claim is only partly true, since the asymptotic performance is not at par with GAIL (except for HalfCheetah where it performs only marginally better). The method also produces high variance policies across different random seeds which is not desirable.
\item \textbf{Random Expert Distillation:} \citet{red} is used in SAIL and it does not use adversarial training to learn a reward function. This method is also more sample efficient than GAIL, but that is only in the first few environment interactions and the peak performance is not as good as GAIL. Our results are consistent with the results reported in \citet{sail}.
\end{itemize}
We use the code provided by \citet{a2ccode} for implementing all baselines. For all experiments, we use a shared value and policy networks, which is an MLP with 2 hidden layers containing 64 hidden units with \textit{tanh} nonlinearities, followed by their individual heads.
For our algorithm, we choose $\alpha_0$ according to the iterations taken for $\alpha_t$ to reach the value $0.5$, which we denote as the `half-life'.
The half-life $H$ is related to the value of $\alpha_0$ as $\alpha_0^{H} = 0.5 \implies \alpha_0 = (0.5)^{\frac{1}{H}}$.
We choose a half-life of $10$ iterations across all experiments, which corresponds to $\alpha_0 \sim 0.933$.
This value was found empirically by running the behavior cloning baseline and setting it equal to half the number of epochs it took for behavior cloning to converge.
All algorithms (except the BC+GAIL baseline) are trained from scratch.
Each algorithm is run across 3 random seeds, as done in \citet{ppo}.
The behavior cloning algorithm is trained only on 70\% of the data, and 30\% is used for validation. For all other experiments, all of the data is used.
Note that although our loss contains a behavior cloning term, it does not require any validation data.
The final performance of each method is evaluated by taking an average of 20 episodes for each seed.
The final performance is shown in Table \ref{tab:bcgail} and the reward curves are shown in Figure \ref{fig:bcgail}.
\begin{figure*}[htpb]
    \centering
    \includegraphics[width=0.29\textwidth]{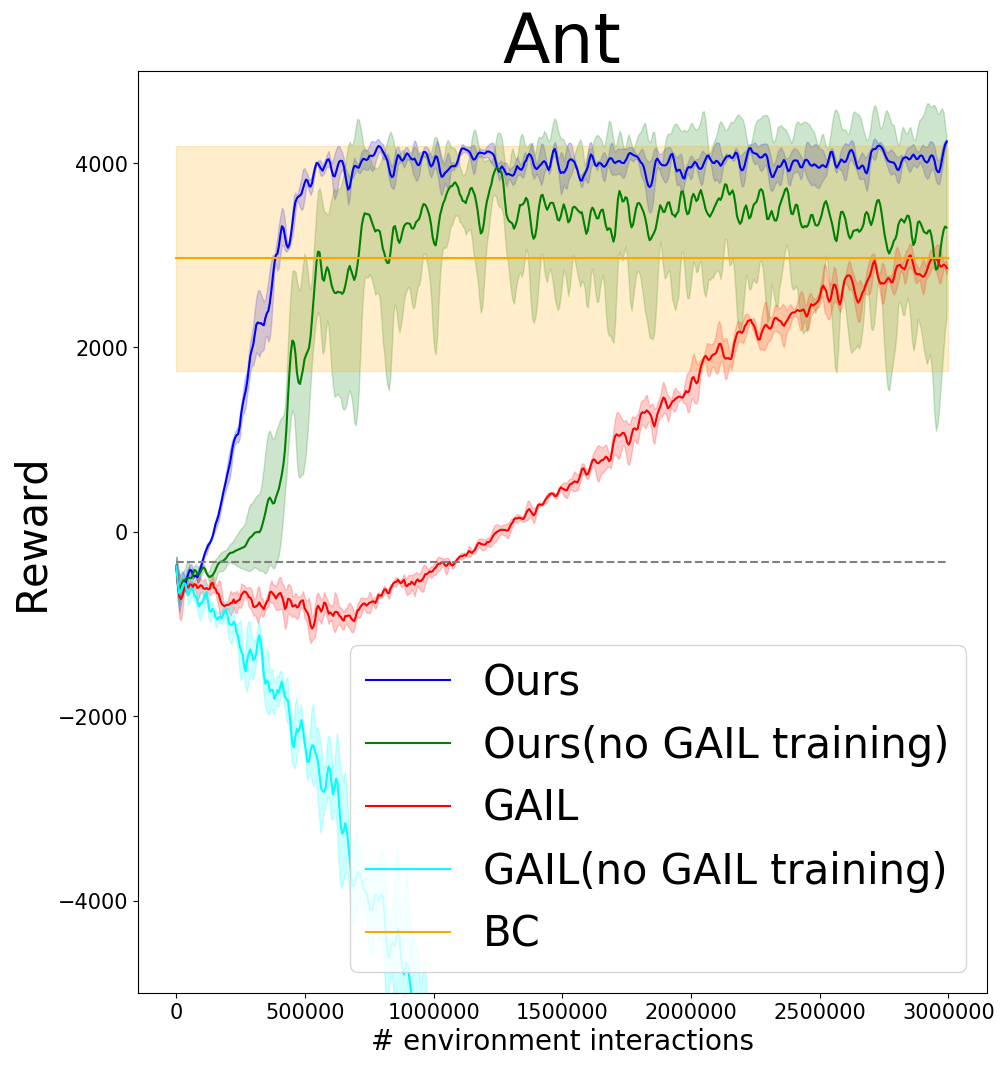}
    \includegraphics[width=0.29\textwidth]{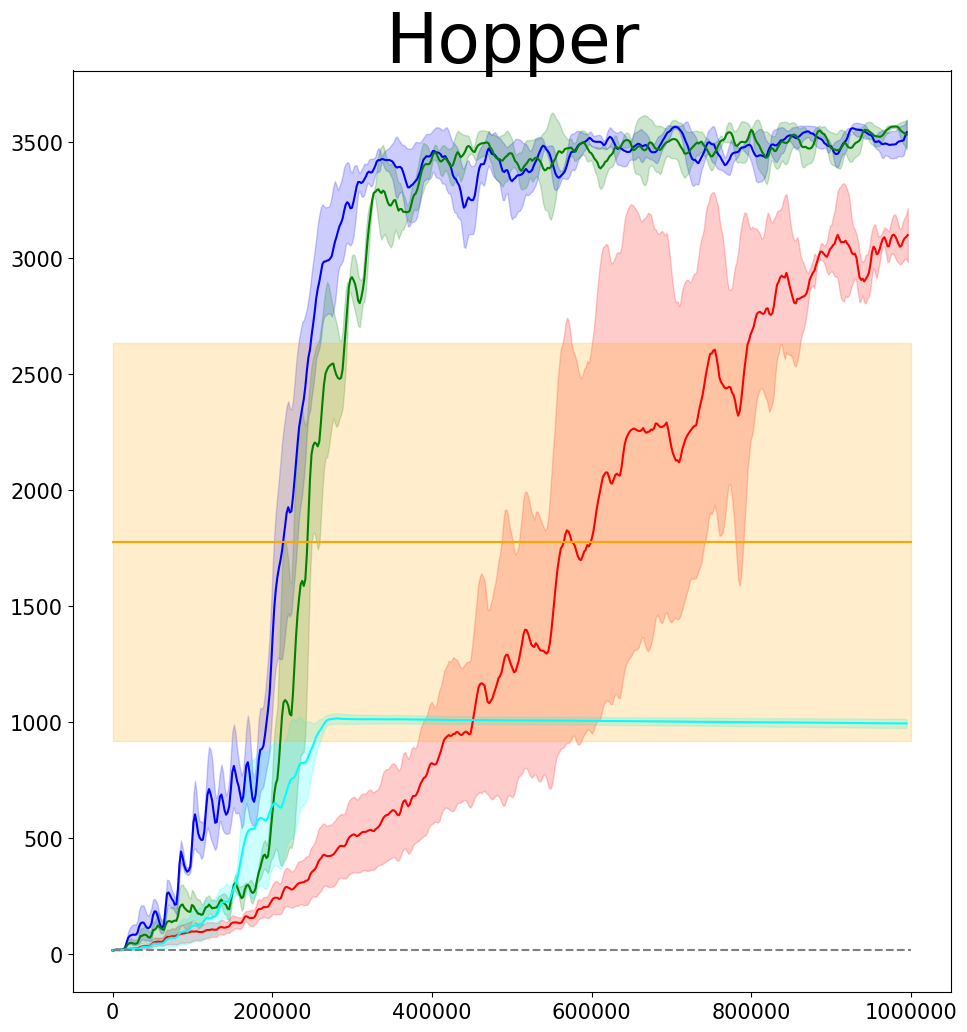}
    \includegraphics[width=0.29\textwidth]{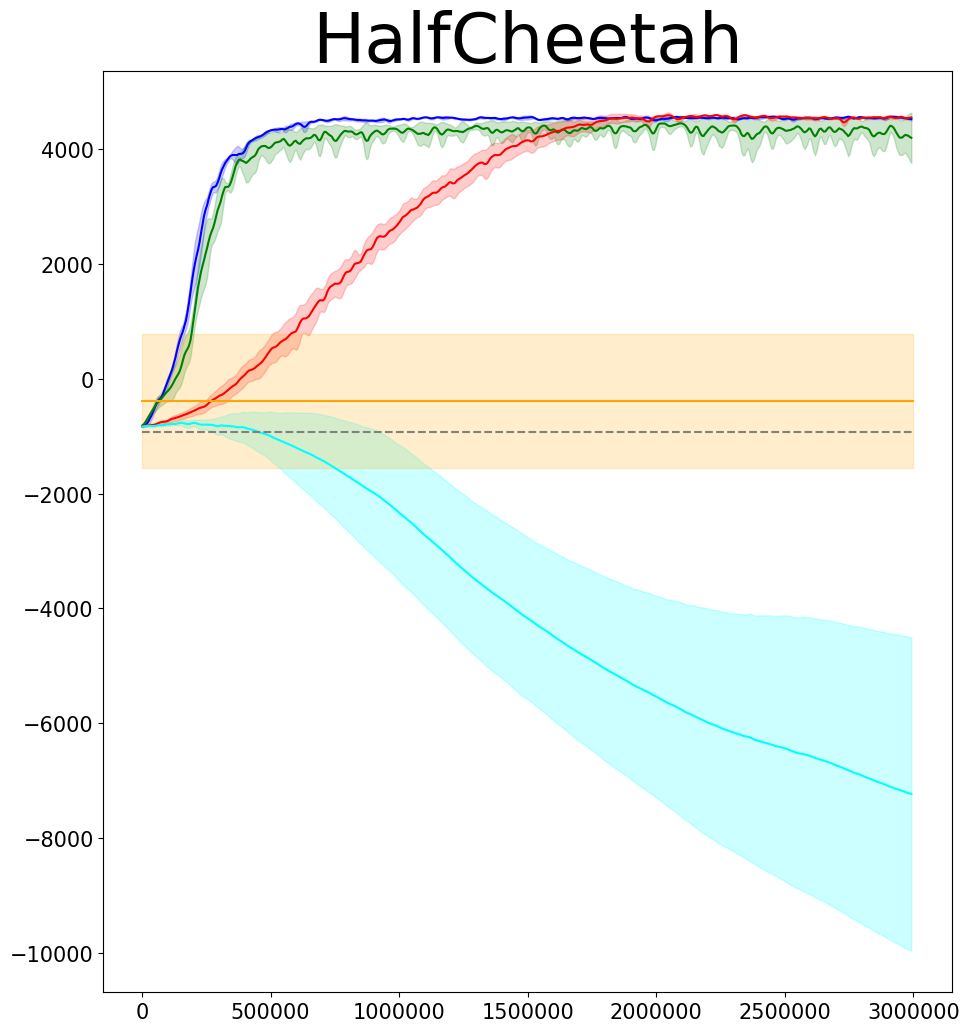}
    \includegraphics[width=0.29\textwidth]{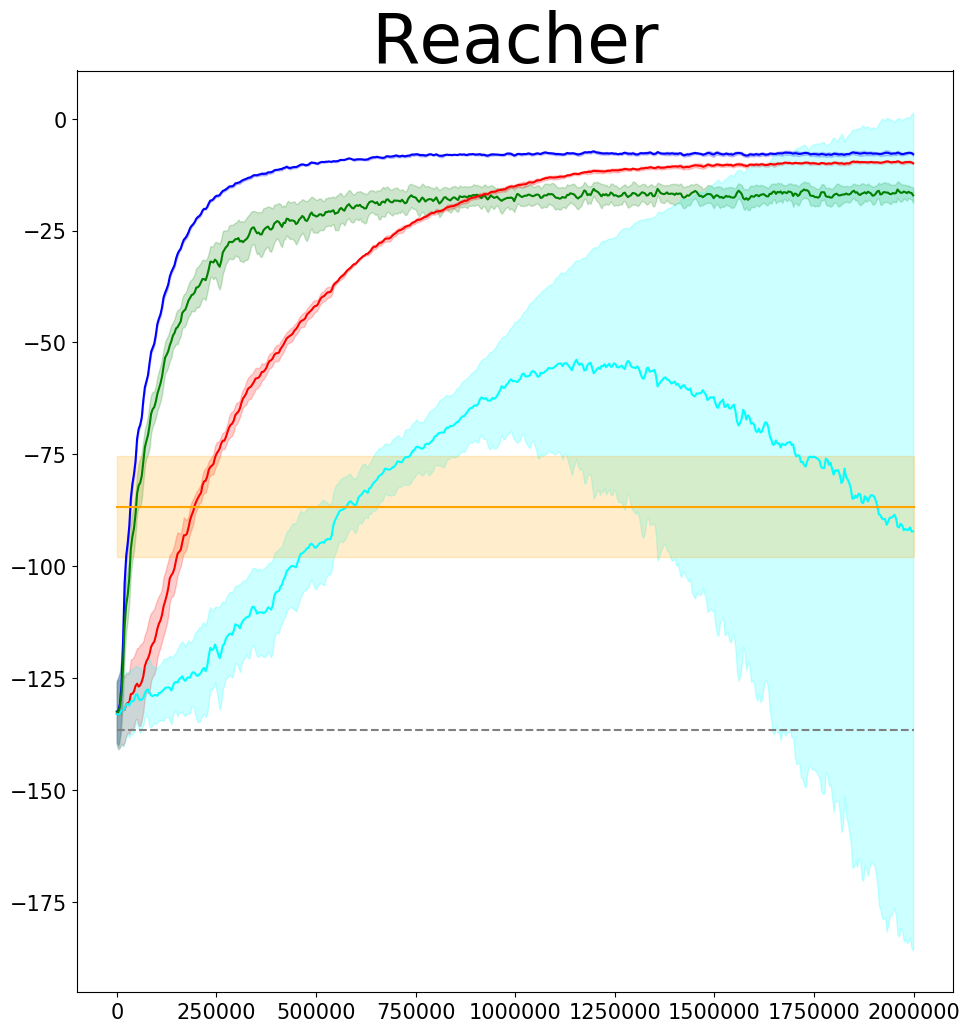}
    \includegraphics[width=0.29\textwidth]{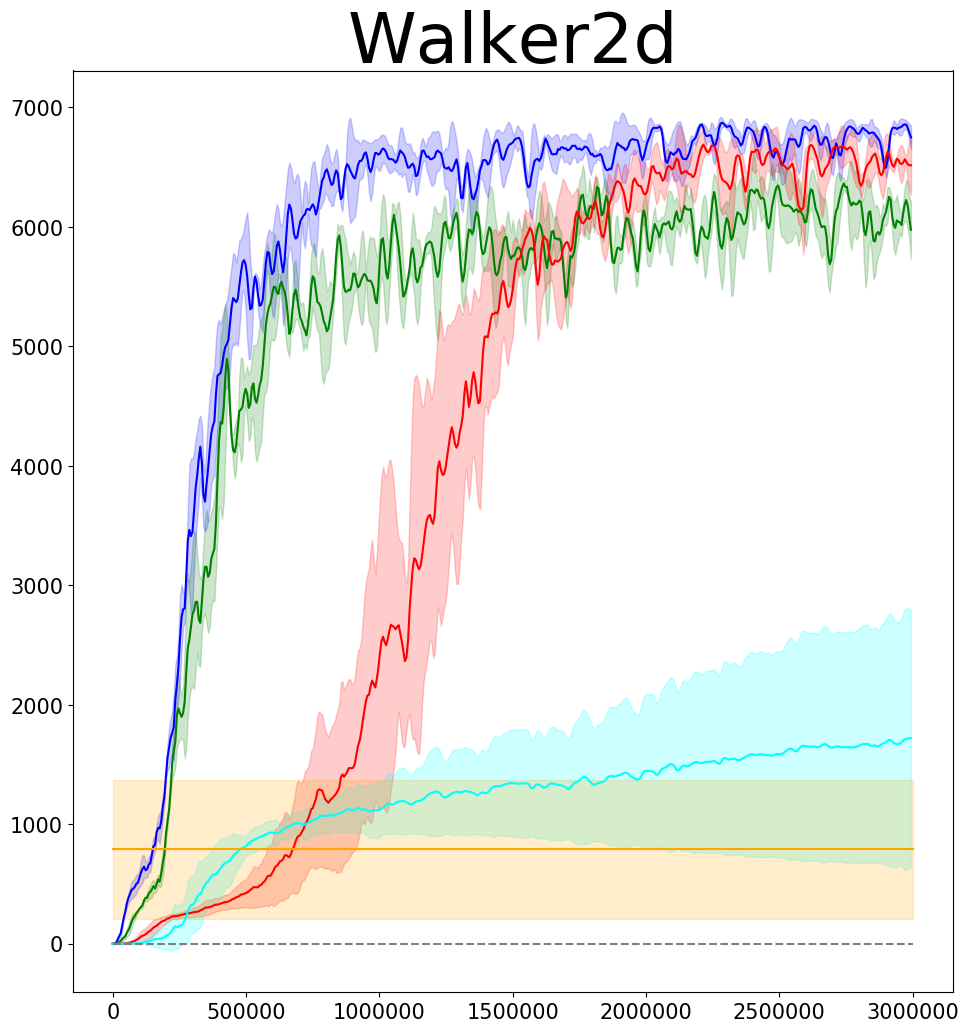}
    \caption{Performance of our method with and without discriminator training. Notice that our method outperforms BC even with random rewards from the discriminator, which shows that adding a temporal dependency in behavior cloning improves performance significantly.}
    \label{fig:bcnogail}
\end{figure*}

Our method performs consistently across all the environments, whereas GAIL is very slow and behaviour cloning never reaches the best performance.
SAIL seems to be outperforming GAIL initially, however, GAIL catches up and has better asymptotic performance than SAIL.
SAIL also has a very high variance compared to other methods, potentially due to amplification of the variance of the two rewards used in their algorithm.
The authors report the best agent performance across all seeds, which obscures the overall stability of the method.
RED performs suboptimally because there is no feedback received from the agent to the reward function for adjusting its reward.
Our method is very sample efficient as it learns much faster than GAIL, and in a lot of cases, converges to a slightly higher reward than GAIL.

\subsection{Effect of temporal dependencies on Behavior Cloning}
\label{bctemp}
\begin{wrapfigure}[40]{R}{0.5\textwidth}
    \centering
    \includegraphics[width=0.24\textwidth]{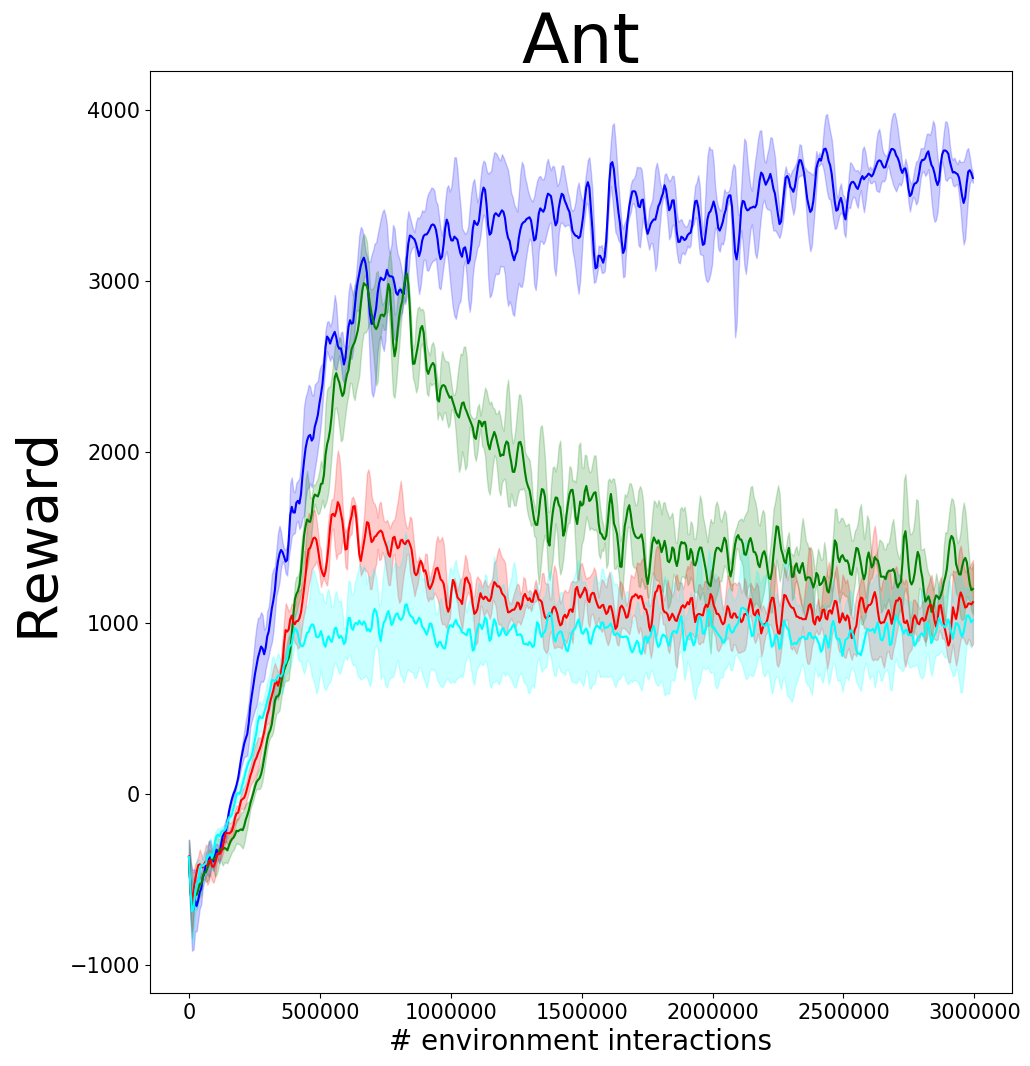}
    \includegraphics[width=0.24\textwidth]{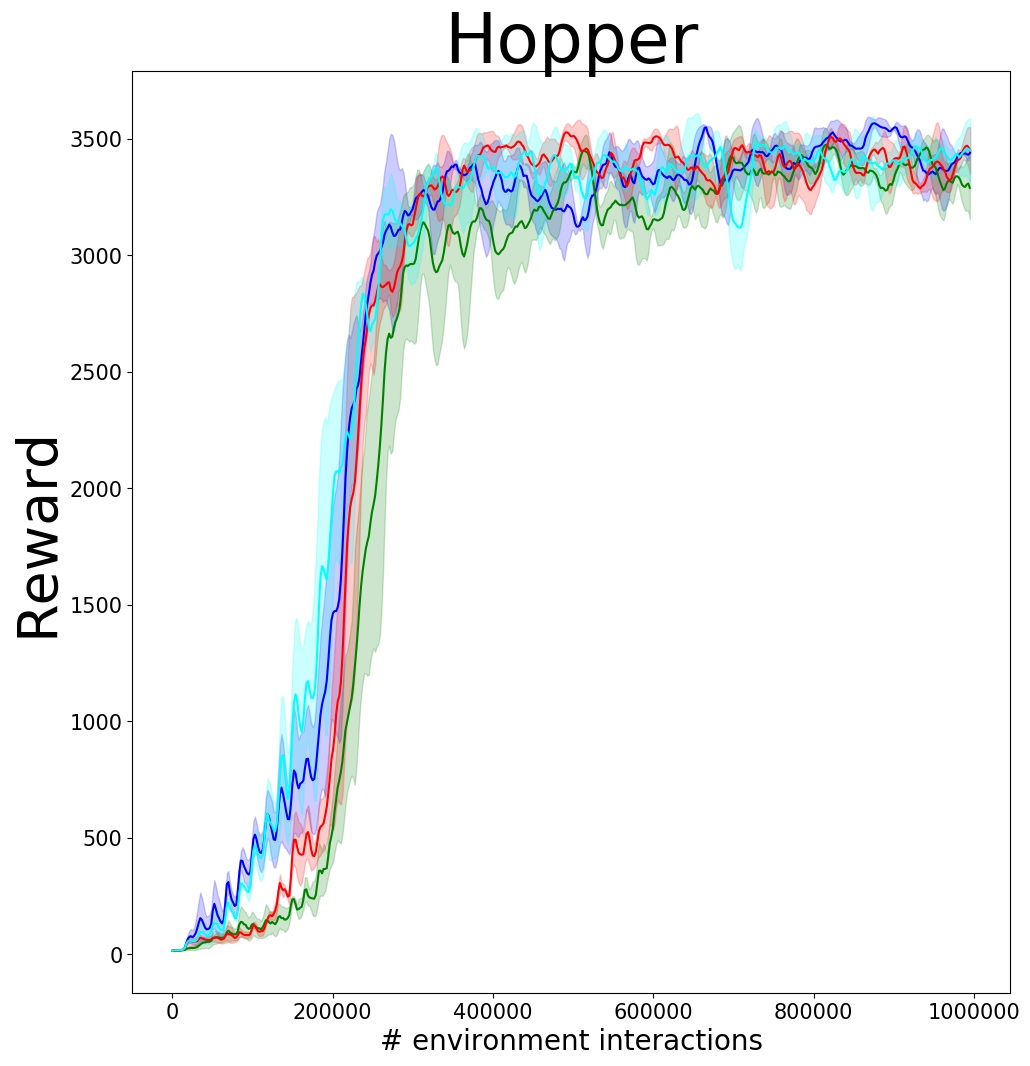}
    \includegraphics[width=0.24\textwidth]{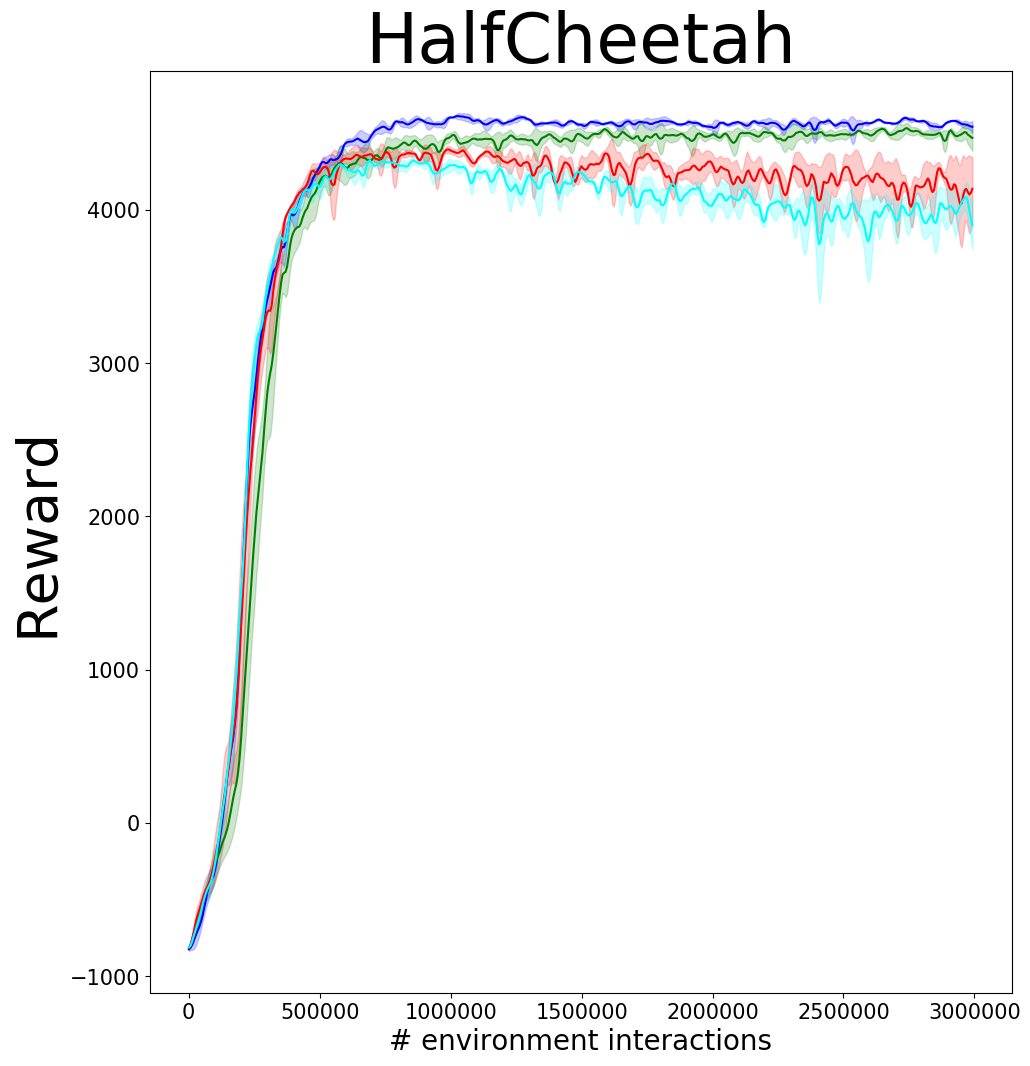}
    \includegraphics[width=0.24\textwidth]{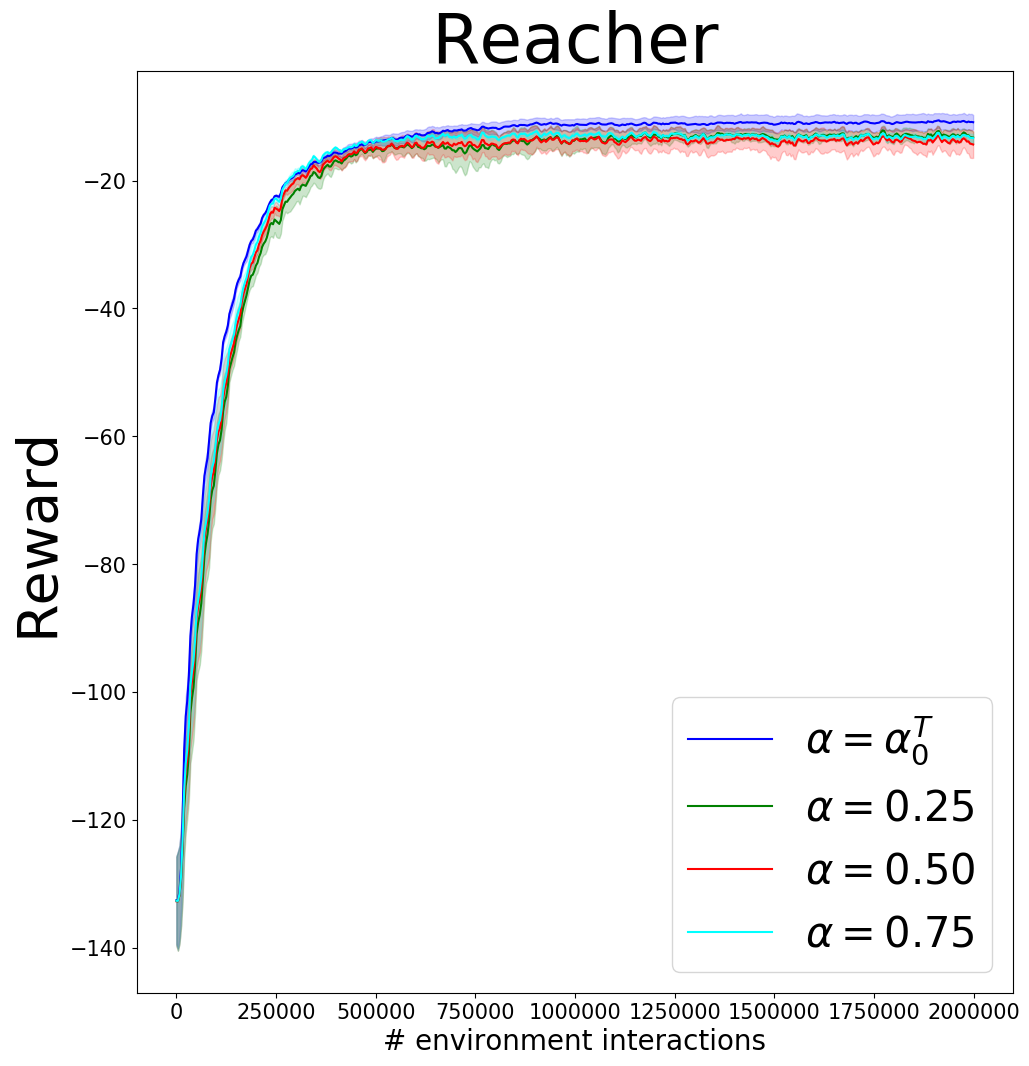}
    \includegraphics[width=0.24\textwidth]{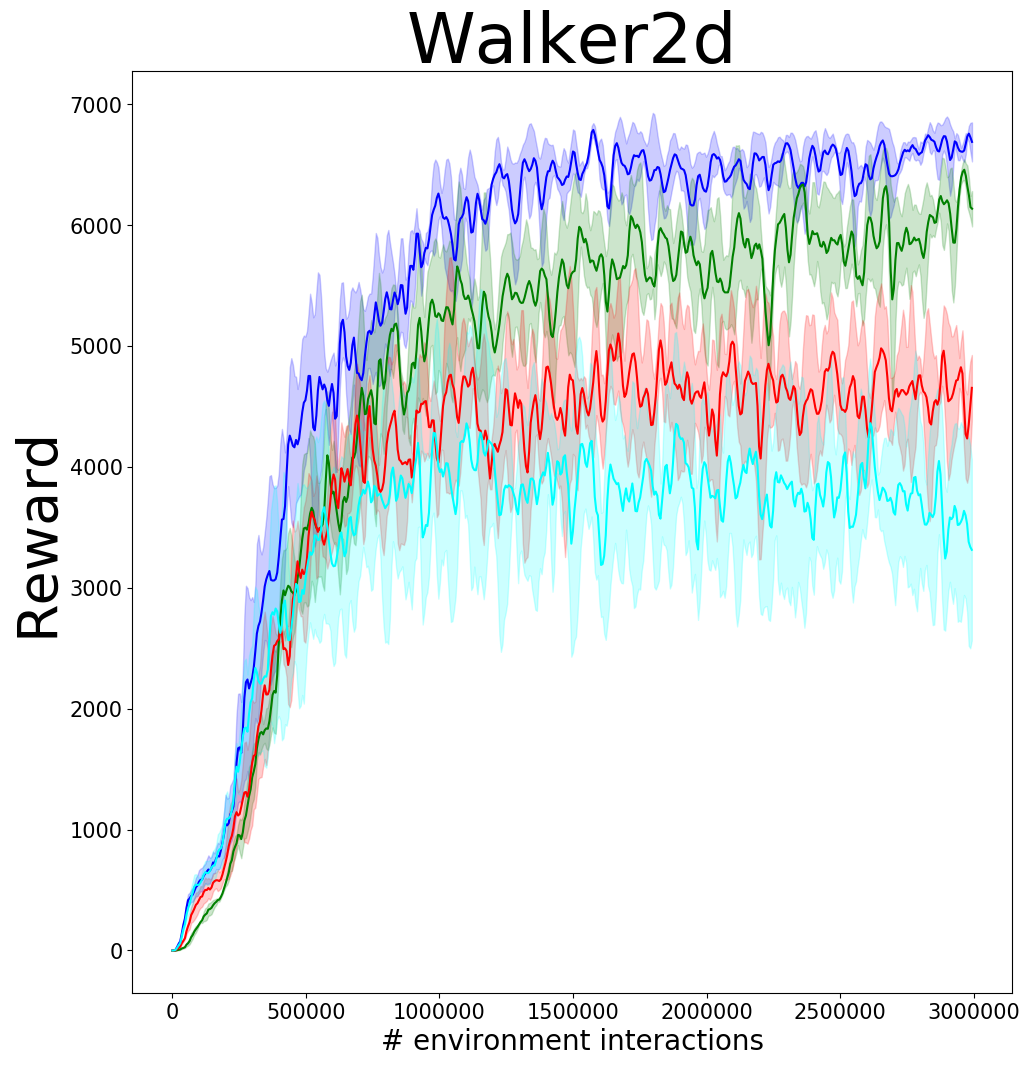}
    \caption{Performance of our method with and without annealing the tradeoff parameter. Notice that the final performance decreases with increasing value of $\alpha$ because the advantage term due to BC contributes in addition to the advantage of RL term, thus rendering the agent prone to overfitting. Our method reaches the best asymptotic performance and is more sample efficient than its constant $\alpha$ variations.}
    \label{fig:bcalphgail}
\end{wrapfigure}
In Section \ref{method} we hypothesized that behavior cloning fails most likely due to miscalibrated actions at out-of-distribution states.
To analyse this effect we train agents with the our method, but we do not train the discriminator.
The second term $A_{\omega, \phi}$ will not provide any useful signal since the discriminator is not trained.
Therefore, the only useful signal can come from the behavior cloning term, and the GAIL term ensures that policy is trained with these uninformative advantage terms.
Since the GAIL term is uninformative, we cannot anneal the value of $\alpha$, otherwise the random rewards can interfere with the agent's learning.
Therefore, we fix $\alpha = 0.5$ for this experiment.
Figure \ref{fig:bcnogail} contains the performance of using our method with random rewards from the discriminator.
The sample efficiency is better than GAIL, and the asymptotic performance is better than behavior cloning, which suggests that behavior cloning can be a powerful candidate for imitation learning.
The untrained GAIL is also plotted to show the effect of potential reward bias that may occur.
We observe that reward bias does not contribute to the task reward, with the exception of Hopper.
This baseline is better than behavior cloning because the agent learns to output more random actions at the states which are outside the expert distribution because the uninformative advantage function does not prefer any action over the other in those states.
The agent learns to perform the expert action at the states that are in the distribution of the expert.
In contrast, behavior cloning never encounters out-of-distribution states during its training, and might output less random actions in those states due to network miscalibration \citet{calibration}.
The positive reward function offers bias only in encouraging survival and not necessarily in achieving a high task reward.

\subsection{Effect of annealing}


Next, we show the effect of annealing versus a fixed value of $\alpha$ on the final performance in all the MuJoCo tasks.
To show that overfitting might be an issue, we limit the number of expert trajectories available to the imitation learning algorithms.
Specifically, we use only $1$ full expert trajectory for learning.
The reward curves in Figure \ref{fig:bcalphgail} demonstrate that as the value of  $\alpha$ increases, the agent learns to imitate faster, but the asymptotic performance does not reach as far as the agents with a lower value of  $\alpha$.
This is the speed versus performance tradeoff associated with  $\alpha$.
To have the best of both worlds,  $\alpha$ is annealed from a high value which promotes faster learning, and is annealed to 0 for better peak performance.
The graphs show that the agent with annealing learns faster \textit{and} achieves the best performance, especially in Ant and Walker environments.

\subsection{Imitation learning with RL in Grid World environments}
Imitation learning can also be used to provide a signal in addition to the environment rewards to enable faster learning (\citet{brys2015rl}, \citet{nair2018overcoming}, \citet{hester2017deep}), especially if the environment rewards are sparse.
We use a gridworld environment as used in \citet{sukhbaatar2018learning}.
We evaluate on the ``Key-Door'' task, where the grid is divided into two rooms.
The agent has to pick up a key, open the locked door and move to the goal location in the other room.
The wall, key, door, goal location and agent are initialized at a different location every time, making the task harder.
The agent recieves a reward of $1$ for reaching the goal location, and $0$ otherwise.
This sparse reward does not provide information about the preconditions that need to be satisfied to reach the goal, i.e. picking the key and unlocking the door.
The expert trajectories, however, contain this information and the agent is rewarded in the short-term horizon for imitating these behaviors.
Therefore, imitation learning can be an extra learning signal for faster learning.

We use the code provided by \citet{sukhbaatar2018learning} and extend it for training the imitation learning algorithms.
The input is a $H\times W$ grid corresponding to the object present in each grid cell.
To prevent the problems of reward bias in this setting \citet{rewardbias}, we follow the work of \citet{infogail} and opt to use a Wasserstein GAN (\citet{wgan}) framework which provides a reward that can be positive or negative.
In addition, we use the REINFORCE algorithm as another baseline which uses the sparse reward.
Experiments show that imitation learning can significantly boost learning compared to a sparse reward signal.
The expert trajectories are collected from an A* agent.
We test with grid sizes of 8, 10, and 12 to analyse the effect of progressively tougher environments.
We use a total of 200, 350, and 500 expert trajectories for grid sizes 8, 10, 12 respectively.
Since expert trajectories are small, there are a lot of unvisited states, and behavior cloning is expected to perform very suboptimally.
\begin{figure*}[htpb]
    \centering
    \includegraphics[width=0.28\textwidth]{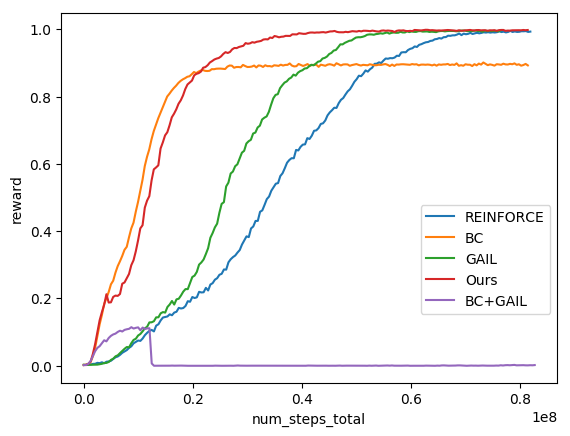}
    \includegraphics[width=0.28\textwidth]{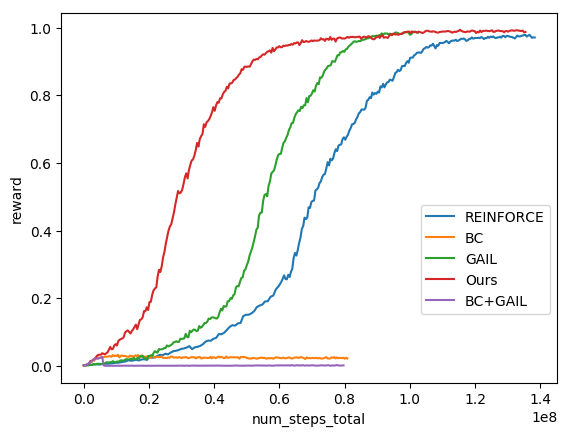}
    \includegraphics[width=0.28\textwidth]{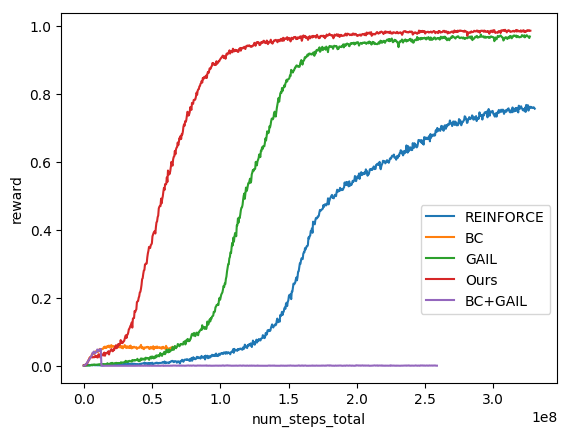}
    \caption{Performance in GridWorld environments. From left to right, the reward curves are for $8\times 8, 10\times 10, 12 \times 12$ grids respectively. Note that our method performs better than GAIL consistently across grid sizes. Behavior cloning is implemented by setting $\alpha = 1$ within our framework for ease of implementation (therefore the reward curve for behavior cloning).}
    \label{fig:gridworld}
\end{figure*}
Figure \ref{fig:gridworld} provides a comparison of all baselines.
Behavior cloning performs suboptimally for all grid sizes, and its performance worsens with increasing grid size.
Policy gradient learns slowly owing to a sparse reward function.
In the case of grid size 12, REINFORCE only reaches about 70\% of the performance of GAIL and our method after 30M steps.
The performance of the BC+GAIL baseline drops to 0 after the behavior cloning stage, and never recovers, showing similar effects to that of the MuJoCo experiments.
This suggests that pretraining with behavior cloning is not a good option across environments and different RL implementations.
However, our method reaches the same performance much faster than GAIL.

\subsection{Imitation Learning in Image-based Environments}
To demonstrate the effectiveness of incorporating the behavior cloning term into GAIL, we compare the variety of methods on Car Racing, a continous control task.
In this environment,the agent must learn to keep a car on track from a top view of the car.
We train an expert using PPO and collect 20 trajectories from randomly selected tracks.
To train all agents, we concatenate the last four frames as the state as a single frame does not encode time dependent variables like velocity and acceleration.
This setting is the same as \citet{atari} for training agents without learning recurrent networks.
\begin{wrapfigure}[22]{R}{0.35\textwidth}
    \begin{tabular}{|c|c|}  \hline
           & \textbf{Score} \\ \hline
        \textbf{Random}  & $-75.01 \pm 4.10$ \\ \hline
        \textbf{BC}  & $695.36 \pm 97.63$ \\ \hline
        \textbf{GAIL}  & $419.82 \pm 198.61$ \\ \hline
        \textbf{BC+GAIL}  & $594.86 \pm 263.12$ \\ \hline
        \textbf{Ours}  & $\boldsymbol{732.55 \pm 45.73}$ \\ \hline
        \textbf{Expert}  & $740.42 \pm 86.36$ \\ \hline
    \end{tabular}
    \includegraphics[width=0.35\textwidth]{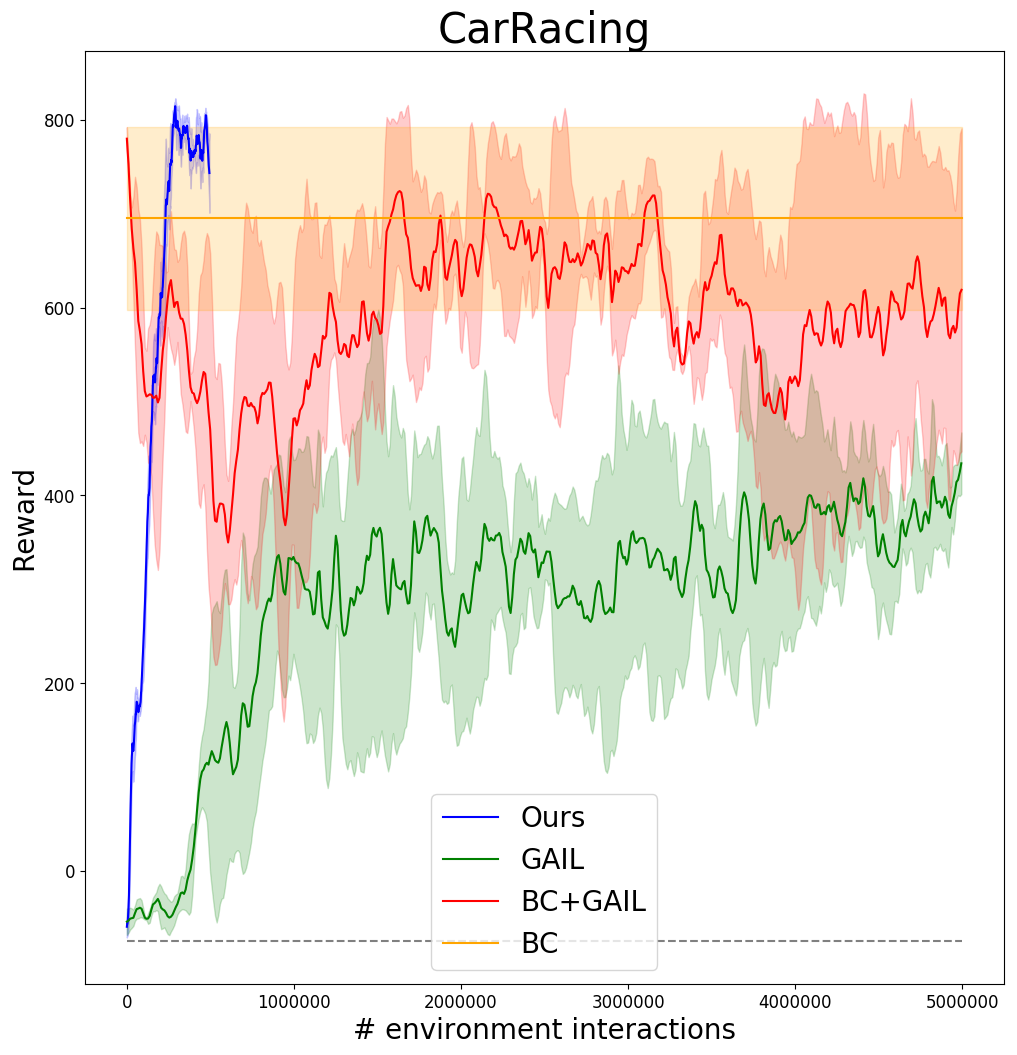}
    \caption{Performance on Car Racing environment}
    \label{fig:carracing}
    \label{tab:carracing}
\end{wrapfigure}

The results in Table \ref{tab:carracing} demonstrate that our method learns a policy that recovers a near-optimal policy from the expert trajectories.
GAIL tends to fail because the agent has to learn image features from a noisy reward signal coming from the discriminator, which is significantly harder than MuJoCo due to curse of dimensionality.
Since this environment can be solved greedily in most parts without solving the credit assignment problem, behavior cloning is already a very strong baseline \citet{sqil}.
Our method uses this aspect of behavior cloning and is able to recover a policy.
The reward curve in Figure \ref{fig:carracing} shows that our method is at least 10x sample efficient than GAIL without resorting to any off-policy schemes.
A model pretrained with behavior cloning starts off with a very good score, but the noisy GAIL reward interferes with its performance and this baseline also performs suboptimally.
Our method is relatively stable and spares the usage of a lot of environment interactions.

\section{Conclusion}

As demonstrated, we show that our method provides stability in adversarial imitation learning, especially in low dimensional tasks with less expert data and in high dimensional tasks where adversarial learning methods are unstable.
In both low and high dimensional tasks, we observe that behavior cloning learns in a few iterations but performs suboptimally.
GAIL learns much slower than behavior cloning but reaches optimal performance.
Pre-training with BC collapses due to problems with initialization and warm starting, and the policy does not converge to peak performance during GAIL training.
Our method combines the best of both worlds by maintaining optimal asymptotic performance, and learns in upto an order of magnitude faster than GAIL.
Experiments on different types of environments demonstrate that our method improves substantially on GAIL by covering up for its one major weakness, its sample efficiency, without compromising on its stability or ease of implementation.

\section{Acknowledgements}
This work was supported by DARPA Award HR001120C0036.

\clearpage
\bibliography{corl}  

\begin{thebibliography}{28}
\providecommand{\natexlab}[1]{#1}
\providecommand{\url}[1]{\texttt{#1}}
\expandafter\ifx\csname urlstyle\endcsname\relax
  \providecommand{\doi}[1]{doi: #1}\else
  \providecommand{\doi}{doi: \begingroup \urlstyle{rm}\Url}\fi

\bibitem[Fujimoto et~al.(2018)Fujimoto, Meger, and Precup]{trulyoffpolicy}
S.~Fujimoto, D.~Meger, and D.~Precup.
\newblock Off-policy deep reinforcement learning without exploration.
\newblock \emph{arXiv preprint arXiv:1812.02900}, 2018.

\bibitem[Ross and Bagnell(2010)]{forwardtraining}
S.~Ross and D.~Bagnell.
\newblock Efficient reductions for imitation learning.
\newblock In \emph{Proceedings of the thirteenth international conference on
  artificial intelligence and statistics}, pages 661--668, 2010.

\bibitem[Daum{\'e} et~al.(2009)Daum{\'e}, Langford, and Marcu]{searn}
H.~Daum{\'e}, J.~Langford, and D.~Marcu.
\newblock Search-based structured prediction.
\newblock \emph{Machine learning}, 75\penalty0 (3):\penalty0 297--325, 2009.

\bibitem[Ross et~al.(2011)Ross, Gordon, and Bagnell]{dagger}
S.~Ross, G.~Gordon, and D.~Bagnell.
\newblock A reduction of imitation learning and structured prediction to
  no-regret online learning.
\newblock In \emph{Proceedings of the fourteenth international conference on
  artificial intelligence and statistics}, pages 627--635, 2011.

\bibitem[Ho and Ermon(2016)]{ho2016generative}
J.~Ho and S.~Ermon.
\newblock Generative adversarial imitation learning.
\newblock In \emph{Advances in neural information processing systems}, pages
  4565--4573, 2016.

\bibitem[Li et~al.(2017)Li, Song, and Ermon]{infogail}
Y.~Li, J.~Song, and S.~Ermon.
\newblock Infogail: Interpretable imitation learning from visual
  demonstrations.
\newblock In I.~Guyon, U.~V. Luxburg, S.~Bengio, H.~Wallach, R.~Fergus,
  S.~Vishwanathan, and R.~Garnett, editors, \emph{Advances in Neural
  Information Processing Systems 30}, pages 3812--3822. Curran Associates,
  Inc., 2017.
\newblock URL
  \url{http://papers.nips.cc/paper/6971-infogail-interpretable-imitation-learning-from-visual-demonstrations.pdf}.

\bibitem[Fu et~al.(2017)Fu, Luo, and Levine]{airl}
J.~Fu, K.~Luo, and S.~Levine.
\newblock Learning robust rewards with adversarial inverse reinforcement
  learning.
\newblock \emph{arXiv preprint arXiv:1710.11248}, 2017.

\bibitem[Sasaki et~al.(2019)Sasaki, Yohira, and Kawaguchi]{sasaki2018sample}
F.~Sasaki, T.~Yohira, and A.~Kawaguchi.
\newblock Sample efficient imitation learning for continuous control.
\newblock In \emph{International Conference on Learning Representations}, 2019.
\newblock URL \url{https://openreview.net/forum?id=BkN5UoAqF7}.

\bibitem[He et~al.(2019)He, Girshick, and Doll{\'a}r]{he2019rethinking}
K.~He, R.~Girshick, and P.~Doll{\'a}r.
\newblock Rethinking imagenet pre-training.
\newblock In \emph{Proceedings of the IEEE International Conference on Computer
  Vision}, pages 4918--4927, 2019.

\bibitem[Ash and Adams(2019)]{ash2019difficulty}
J.~T. Ash and R.~P. Adams.
\newblock On the difficulty of warm-starting neural network training.
\newblock \emph{arXiv preprint arXiv:1910.08475}, 2019.

\bibitem[Pan et~al.(2019)Pan, Niu, Li, Dou, and Jiang]{PAN2019}
H.~Pan, X.~Niu, R.~Li, Y.~Dou, and H.~Jiang.
\newblock Annealed gradient descent for deep learning.
\newblock \emph{Neurocomputing}, 2019.
\newblock ISSN 0925-2312.
\newblock \doi{https://doi.org/10.1016/j.neucom.2019.11.021}.
\newblock URL
  \url{http://www.sciencedirect.com/science/article/pii/S0925231219315802}.

\bibitem[Fujimoto et~al.(2018)Fujimoto, van Hoof, and
  Meger]{fujimoto2018addressing}
S.~Fujimoto, H.~van Hoof, and D.~Meger.
\newblock Addressing function approximation error in actor-critic methods.
\newblock \emph{arXiv preprint arXiv:1802.09477}, 2018.

\bibitem[Jena and Awate(2019)]{jena}
R.~Jena and S.~P. Awate.
\newblock A bayesian neural net to segment images with uncertainty estimates
  and good calibration.
\newblock In \emph{Information Processing in Medical Imaging}, pages 3--15,
  Cham, 2019. Springer International Publishing.
\newblock ISBN 978-3-030-20351-1.

\bibitem[Wang et~al.(2019{\natexlab{a}})Wang, Ciliberto, Amadori, and
  Demiris]{sail}
R.~Wang, C.~Ciliberto, P.~V. Amadori, and Y.~Demiris.
\newblock Random expert distillation: Imitation learning via expert policy
  support estimation.
\newblock \emph{CoRR}, abs/1905.06750, 2019{\natexlab{a}}.
\newblock URL \url{http://arxiv.org/abs/1905.06750}.

\bibitem[Wang et~al.(2019{\natexlab{b}})Wang, Ciliberto, Amadori, and
  Demiris]{red}
R.~Wang, C.~Ciliberto, P.~Amadori, and Y.~Demiris.
\newblock Random expert distillation: Imitation learning via expert policy
  support estimation.
\newblock \emph{arXiv preprint arXiv:1905.06750}, 2019{\natexlab{b}}.

\bibitem[Kostrikov(2018)]{a2ccode}
I.~Kostrikov.
\newblock Pytorch implementations of reinforcement learning algorithms.
\newblock \url{https://github.com/ikostrikov/pytorch-a2c-ppo-acktr-gail}, 2018.

\bibitem[Schulman et~al.(2017)Schulman, Wolski, Dhariwal, Radford, and
  Klimov]{ppo}
J.~Schulman, F.~Wolski, P.~Dhariwal, A.~Radford, and O.~Klimov.
\newblock Proximal policy optimization algorithms.
\newblock \emph{arXiv preprint arXiv:1707.06347}, 2017.

\bibitem[Guo et~al.(2017)Guo, Pleiss, Sun, and Weinberger]{calibration}
C.~Guo, G.~Pleiss, Y.~Sun, and K.~Q. Weinberger.
\newblock On calibration of modern neural networks, 2017.

\bibitem[Brys et~al.(2015)Brys, Harutyunyan, Suay, Chernova, Taylor, and
  Now{\'e}]{brys2015rl}
T.~Brys, A.~Harutyunyan, H.~B. Suay, S.~Chernova, M.~E. Taylor, and
  A.~Now{\'e}.
\newblock Reinforcement learning from demonstration through shaping.
\newblock In \emph{Twenty-fourth international joint conference on artificial
  intelligence}, 2015.

\bibitem[Nair et~al.(2018)Nair, McGrew, Andrychowicz, Zaremba, and
  Abbeel]{nair2018overcoming}
A.~Nair, B.~McGrew, M.~Andrychowicz, W.~Zaremba, and P.~Abbeel.
\newblock Overcoming exploration in reinforcement learning with demonstrations.
\newblock In \emph{2018 IEEE International Conference on Robotics and
  Automation (ICRA)}, pages 6292--6299. IEEE, 2018.

\bibitem[Hester et~al.(2017)Hester, Vecerik, Pietquin, Lanctot, Schaul, Piot,
  Horgan, Quan, Sendonaris, Dulac-Arnold, et~al.]{hester2017deep}
T.~Hester, M.~Vecerik, O.~Pietquin, M.~Lanctot, T.~Schaul, B.~Piot, D.~Horgan,
  J.~Quan, A.~Sendonaris, G.~Dulac-Arnold, et~al.
\newblock Deep q-learning from demonstrations.
\newblock \emph{arXiv preprint arXiv:1704.03732}, 2017.

\bibitem[Sukhbaatar et~al.(2018)Sukhbaatar, Denton, Szlam, and
  Fergus]{sukhbaatar2018learning}
S.~Sukhbaatar, E.~Denton, A.~Szlam, and R.~Fergus.
\newblock Learning goal embeddings via self-play for hierarchical reinforcement
  learning.
\newblock \emph{arXiv preprint arXiv:1811.09083}, 2018.

\bibitem[Kostrikov et~al.(2018)Kostrikov, Agrawal, Dwibedi, Levine, and
  Tompson]{rewardbias}
I.~Kostrikov, K.~K. Agrawal, D.~Dwibedi, S.~Levine, and J.~Tompson.
\newblock Discriminator-actor-critic: Addressing sample inefficiency and reward
  bias in adversarial imitation learning.
\newblock \emph{arXiv preprint arXiv:1809.02925}, 2018.

\bibitem[Arjovsky et~al.(2017)Arjovsky, Chintala, and Bottou]{wgan}
M.~Arjovsky, S.~Chintala, and L.~Bottou.
\newblock Wasserstein gan.
\newblock \emph{arXiv preprint arXiv:1701.07875}, 2017.

\bibitem[Mnih et~al.(2013)Mnih, Kavukcuoglu, Silver, Graves, Antonoglou,
  Wierstra, and Riedmiller]{atari}
V.~Mnih, K.~Kavukcuoglu, D.~Silver, A.~Graves, I.~Antonoglou, D.~Wierstra, and
  M.~Riedmiller.
\newblock Playing atari with deep reinforcement learning.
\newblock \emph{arXiv preprint arXiv:1312.5602}, 2013.

\bibitem[Reddy et~al.(2019)Reddy, Dragan, and Levine]{sqil}
S.~Reddy, A.~D. Dragan, and S.~Levine.
\newblock Sqil: imitation learning via regularized behavioral cloning.
\newblock \emph{arXiv preprint arXiv:1905.11108}, 2019.

\bibitem[maz()]{mazebase}
Mazebase: A sandbox for learning from games.
\newblock \url{https://github.com/facebookarchive/MazeBase}.

\bibitem[Brockman et~al.(2016)Brockman, Cheung, Pettersson, Schneider,
  Schulman, Tang, and Zaremba]{gym}
G.~Brockman, V.~Cheung, L.~Pettersson, J.~Schneider, J.~Schulman, J.~Tang, and
  W.~Zaremba.
\newblock Openai gym, 2016.

\end{thebibliography}

\clearpage
\section{Appendix}
In this section, we discuss how our work can be extended to real robots.

\subsection{Extending to real robots}
Our method can be easily extended to real robots since we extend GAIL with behavior cloning, and both methods are used in real robots.
We use on-policy optimization which can been applied in real robots as well.
To test that our method is not sensitive to any particular environment, we test our method on three different simulators.
The first simulator comes from MuJoCo, which is a physics engine and the environments we use (Hopper, HalfCheetah, Ant, Walker2d, Reacher) are analogous to physical robots.
The second simulator is MazeBase \citet{mazebase} which is a sandbox containing various mazes with different tasks to solve.
The results of experiments on this simulator can be used to verify the performance of our method on environments where planning and navigation is required.
The third simulator we use is the CarRacing environment in OpenAI gym \citet{gym}.
This environment is a simplified version of a car driving setup, where the actions also correspond to steer, gas and brake, analogous to a real-life car driving scenario.
The environments in all 3 scenarios have different input formats, ranging from state information to raw pixels.
The environments range from requiring short and long term planning to solve the tasks (for example, the CarRacing environment needs more short term planning than the Mazebase environment).
Our method outperforms baselines in all these scenarios, and hence it would extend to real robots as well with minimal effort.

\end{document}